\documentclass{tlp}
\usepackage{amsmath}
\usepackage{url}
\usepackage{picinpar}
\usepackage{epsfig}

\newtheorem{example}{Example} % [section]

\begin{document}

\long\def\comment#1{}

\setlength{\intextsep}{1pt}
\setlength{\floatsep}{1pt}
\setlength{\textfloatsep}{1pt}

\def\causes{\: \mathbf{causes}\:}
\def\broadcast{\: \mathbf{broadcasts} \:}
\def\inform{\: \mathbf{informs} \: }
\def\oof{\: \mathbf{of} \:\: }
\def\ask{\: \mathbf{ask}}
\def\retract{\: \mathbf{retract}}
\def\determines{\: \mathbf{determines} \:}
\def\executable{\: \mathbf{executable\_if} \:}
\def\iif{\: \mathbf{if} \:}
\def\initially{\: \mathbf{initially} \:}
\def\after{\: \mathbf{after} \:}
\def\naf{\mathtt{not }}

\def\announcesk{\:\:\mathbf{announces}\:\:}
\def\announcesf{\:\:\mathbf{told}\:\:}
\def\to{\:\: \mathbf{to} \:\: }
\def\trysense{\:\: \mathbf{senses} \:\: }

% logic programming and action language macros
\newcommand{\A}{\ensuremath{\mathcal{A}}}
\newcommand{\mA}{\ensuremath{\mathbf{m}\mathcal{A}}}
\newcommand{\AL}{\ensuremath{\mathcal{AL}}}
\newcommand{\Reduct}[2]{\ensuremath{{#1}^{#2}}}
\newcommand{\lparrow}{\leftarrow}
\newcommand{\agents}{\ensuremath{\mathcal{A}}}
\newcommand{\fluents}{\ensuremath{\mathcal{F}}}
\newcommand{\announce}[1]{\ensuremath{{#1}\:\textbf{announced}}}
\newcommand{\announcedto}[2]{\ensuremath{{#1}\:\mbox{\textbf{announced to}}\:{#2}}}
\newcommand{\announcedobs}[3]{\ensuremath{{#1}\:\mbox{\textbf{announced to}}\:{#2} \:\mbox{\textbf{observed by}}\:{#3}}}
\newcommand{\sensed}[2]{\ensuremath{{#1}\:\mbox{\textbf{sensed by}}\:{#2}}}
\newcommand{\trysensed}[2]{\ensuremath{\mbox{\textbf{sense}}\:{#1}\:\mbox{\textbf{attempted by}}\:{#2}}}
\newcommand{\sensedobs}[3]{\ensuremath{{#1}\:\mbox{\textbf{sensed by}}\:{#2}\:\mbox{\textbf{observed by}}\:{#3}}}
\newcommand{\trysensedobs}[3]{\ensuremath{\mbox{\textbf{sense}}\:{#1}\:\mbox{\textbf{attempted by}}\:{#2}\:\mbox{\textbf{observed by}}\:{#3}}}
\newcommand{\successor}[2]{\ensuremath{succ(#1,{#2})}}
\newcommand{\situation}[2]{\ensuremath{({#1},{#2})}}
\newcommand{\performedby}{\ensuremath{\:\:\mathbf{performed\_by}\:\:}}
\newcommand{\observedby}{\ensuremath{\:\:\mathbf{observed\_by}\:\:}}

% general math notation macros
\newcommand{\bottom}{\ensuremath{\bot}}
\newcommand{\entails}{\ensuremath{\models}}
\newcommand{\Union}{\ensuremath{\bigcup}}
\newcommand{\intersect}{\ensuremath{\cap}}
\newcommand{\Conj}{\ensuremath{\bigwedge}}
\newcommand{\Disj}{\ensuremath{\bigvee}}

\newcommand{\union}{\ensuremath{\cup}}

\newcommand{\equivl}{\ensuremath{\Leftrightarrow}}
\newcommand{\cat}{\ensuremath{\circ}}
\newcommand{\tbeg}{\ensuremath{\langle}}
\newcommand{\tend}{\ensuremath{\rangle}}
\newcommand{\pair}[2]{\ensuremath{\tbeg #1,#2 \tend}}
\newcommand{\disjoint}[2]{\ensuremath{{#1} \intersect {#2} = \emptyset}}
\newcommand{\arcminus}{\ensuremath{\overset{a}{\ominus}}}
\newcommand{\stateminus}{\ensuremath{\overset{s}{\ominus}}}
\newcommand{\cequiv}{\ensuremath{\overset{c}{\sim}}}
\newcommand{\ckplus}{\ensuremath{\overset{\kappa}{\union}}}
\newcommand{\kplus}[2]{\ensuremath{\uplus^{{#1}}_{{#2}}}}
\newcommand{\disj}{\ensuremath{\vee}}

\def\ck{{\mathbf{C}}}
\def\ek{{\mathbf{E}}}
\def\k{{\mathbf{K}}}
\def\rk{{\mathcal{K}}}
\def\lk{{\mathcal{LK}}}
\def\ma{{\mathcal{MA}}}
\def\s5{{\mathbf{S5}}}
\def\cala{{\cal A}}
\def\calag{{\cal AG}}
\def\calf{{\cal F}}
\def\after{{\: \hbox{\bf after } \:}}

\newcommand{\beli}{\begin{list}{$\bullet$}{\topsep=1pt \parsep=0pt \itemsep=1pt}}
\newcommand{\enli}{\end{list}}

\title[LP and the Logics of Knowledge]{Logic Programming for Finding Models in the Logics of Knowledge and its Applications: A Case Study}

\author[Baral, Gelfond, Pontelli, Son]
{\begin{tabular}{ccc}
C. Baral, G. Gelfond & \hspace{.5cm} & E. Pontelli, T. Son\\
Dept. Computer Science && Dept. Computer Science\\
Arizona State Univ. && New Mexico State Univ.
\end{tabular}}

\pagerange{\pageref{firstpage}--\pageref{lastpage}}
\volume{\textbf{10} (3):}
\jdate{March 2002}
\setcounter{page}{1}
\pubyear{2002}

\maketitle

\label{firstpage}

\begin{abstract}
The logics of knowledge are modal logics that have been shown to be
effective in  representing and reasoning about knowledge in multi-agent 
domains. Relatively few computational frameworks for dealing with
computation of models and useful transformations in logics of knowledge 
(e.g., to support multi-agent planning with knowledge actions and degrees
of visibility) have been proposed. This paper explores the use of logic
programming (LP) to encode interesting forms of logics of knowledge and compute
Kripke models. The LP modeling is expanded with 
useful operators on Kripke structures, to support multi-agent
planning in the presence of both world-altering and knowledge actions. This results
in the first ever implementation of a planner for this  type of complex multi-agent
domains.

\end{abstract}

\section{Introduction}

Modeling real-world planning scenarios that involve multiple agents
has been an active area of research over the years. 
A considerable source of complexity derives from those scenarios where
agents need to reason about knowledge and capabilities of other
agents in order to accomplish their tasks. For example, a gambling 
agent needs to reason about what other players may know about the
game in order to make the next move.
Reasoning about knowledge and capabilities in multi-agent domains differs
significantly from the same problem in single-agent domains. The complexity 
arises from two  sources: \emph{(1)}
the representation of  a planning domain needs to model not only the
state of the world, but also the \emph{knowledge/beliefs} of the
agents; \emph{(2)} the actions performed by an agent may lead to changes
in knowledge and beliefs of different agents, e.g., action like
 announcements, cheating, lying, etc.

Various \emph{modal logics} have been developed for reasoning about knowledge
in multi-agent systems
(see, e.g., \cite{FaginHMV95,Halpern95,vanDitmarschHK07}). The semantics of several
of these logics is provided in terms of \emph{Kripke structures}---where,
intuitively, each Kripke structure captures the knowledge/beliefs of
all the agents.
Naturally, when reasoning about knowledge
of multiple agents in a \emph{dynamic} environment, it is necessary to devise
 operations for 
updating Kripke structures after the occurrence of an action, as suggested
in  \cite{BaltagM04,BenthemEK06,vanDitmarschHK07}. Two important
questions, that have been less frequently considered, are:
\emph{(1)} how can 
one determine the initial model or the initial Kripke structure of 
a theory encoding the knowledge of the agents, and
\emph{(2)} what udpate operations on Kripke structures are 
necessary to  lay the foundations of a high-level multi-agent action language.

The dominant presence of Kripke structures in different formalizations 
of multi-agent domains indicates 
that any system employing modal logic for reasoning and planning 
in multi-agent domains would need 
to implement some operations for the manipulation of Kripke structures. 
The lessons learned in the research 
 in single-agent domains suggest that 
a high-level action language for multi-agent domains could be very 
useful. The development of such high-level language does not only 
help in modeling but also provides a new opportunity for the 
development of planning systems operating on top of this language. 
The complexity of various reasoning problems in modal logics
(e.g., satisfaction, validity, etc.) and the difficulty in updating 
a Kripke structure after the occurrence of an event provide a 
computational challenge for logic programming. It is interesting 
to observe that, although there have been a few implementations of temporal 
logics in ASP (e.g., \cite{HeljankoN03,SonBTM06}), to the best of 
our knowledge, there has been no attempt in using logic
programming for computing 
models of modal logics except in our recent work \cite{BaralGSP10a}.
Our initial experience reveals that ASP-based implementation encounter
severe difficulties, mostly associated to the grounding requirements
of ASP. This also 
drives us to explore alternatives. 

In this paper, we investigate the use of \emph{Prolog} in the 
development of a computational framework for reasoning and planning 
in multi-agent domains. 
The advantage of Prolog lies in that it allows the step-by-step 
examination of  parts of a Kripke structure without the need of 
constructing the complete structure. 
We focus on two initial aspects: computing the initial 
model of a theory of knowledge and manipulating Kripke structures. 
To this end, we will  make use of a high-level action
language for the specification 
of various types of actions in multi-agent domains.
%We will begin with a short review of various 
%logics of knowledge (Sec. \ref{sec2}) and the high-level action language $\mA$
%(Sec. \ref{sec3}). We then present the encoding for computing the 
%initial Kripke structure (Sec. \ref{sec4}), discuss the transition function 
%of $\mA$ and provide the implementation of $\mA$ (Sec. \ref{sec5})
%and conclude (Sec. \ref{sec6}). 

\section{Background: The Logics of Knowledge}
\label{sec2}
In this paper we follow the notation established in \cite{FaginHMV95}.
The modal language $\lk_\cala$ builds on a signature that contains
a collection of propositions $\calf$ (often referred to as \emph{fluents}),
the traditional propositional connectives, and a finite set
of modal operators $\k_i$ for each $i$ in a set $\cala$. We will
occasionally refer to the elements in the set $\cala$ as \emph{agents}, and
the pair $\langle \cala, \calf\rangle$ as a \emph{multi-agent domain}.

$\lk_\cala$ formulae are defined as follows.
A \emph{fluent formulae} is a propositional
formula built using fluents and the standard Boolean
operators $\vee, \wedge, \neg$, etc.
A  \emph{modal formulae}  is \emph{(i)}
a fluent formula, or \emph{(ii)} a formula of the form $\k_i \psi$
where $\psi$ is a modal formula, or \emph{(iii)} a formula of the
form $\psi \vee \phi$, $\psi \wedge \phi$,
or $\neg \psi$, where $\psi$ and $\phi$ are modal formulae.
In addition, given a formula $\psi$ and a non-empty
set $\alpha \subseteq \cala$:
\begin{list}{$\bullet$}{\parsep=0pt \itemsep=0pt \topsep=1pt}
    \item $\ek_\alpha \psi$ denotes the set of formulae $\{\k_i \psi
\mid i \in \alpha\}$.
    \item $\ck_\alpha \psi $ denotes the set of formulae of the
form $\ek^k_\alpha \psi$, where $k \ge 1$ and $\ek^{k+1}_\alpha \psi =
\ek^{k}_\alpha \ek_\alpha \psi$ (and $\ek^1_\alpha \psi = \ek_\alpha \psi$).
\end{list}
An $\lk_\cala$ theory is a set of $\lk_\cala$ formulae.

The semantics of $\lk_\cala$ theories is given by \emph{Kripke
structures.}
A {\em Kripke structure} is a tuple
$(S,\pi,\{\rk_i\}_{i\in\cala})$, where $S$ is a set of state
symbols, $\pi$ is a function that associates an interpretation of
$\calf$ to each state in $S$,
and $\rk_i \subseteq S\times S$ for $i \in \cala$.

Different logic systems for reasoning with a $\lk_\cala$ theory
 have been introduced; these differ in the additional
axioms that the models are required to satisfy, e.g.,
\begin{list}{$\bullet$}{\parsep=0pt \itemsep=0pt \topsep=1pt}
    \item \textbf{P}: all instances of axioms of propositional logic
    \item \textbf{K}: $(\k_i\varphi \wedge \k_i(\varphi \Rightarrow \phi))\Rightarrow \k_i \phi$
    \item \textbf{T}: $\k_i\varphi \Rightarrow \varphi$
    \item \textbf{4}: $\k_i\varphi \Rightarrow \k_i\k_i \varphi$
    \item \textbf{5}: $\neg \k_i\varphi \Rightarrow \k_i \neg \k_i \varphi$
    \item \textbf{D}: $\neg \k_i false$
\end{list}
For example, the following modal logic systems are frequently used  \cite{Halpern97}:
 $\s5$ satisfies all of the
 above axioms with the exception of (\textbf{D}),
 $\mathbf{KD45}$ includes the four
  axioms {\bf K}, {\bf 4}, {\bf 5}, and {\bf D},
  $\mathbf{S4}$ includes the axioms
  {\bf K}, {\bf T}, and {\bf 4}, and
  $\mathbf{T}$ includes the two 
  axioms {\bf K} and {\bf T}.

%%%

%
Given a Kripke structure $M=(S,\pi,\{\rk_i\}_{i\in\cala})$ and a state $s\in S$,
we refer to the pair $(M,s)$ as a \emph{pointed Kripke structure}---and $s$ is
referred to as the \emph{real state}.
The  satisfaction relation between $\lk_\cala$-formulae
and a pointed Kripke structure $(M,s)$ is defined as follows: 
({\em i}) $(M,s) \models \varphi$
if $\varphi$ is a fluent formula and $\pi(s) \models \varphi$;
({\em ii}) $(M,s) \models \k_i \varphi$ if $(M,s') \models \varphi$
for every $s'$ such that $(s,s') \in \rk_i$; and ({\em iii})
$(M,s) \models \neg \varphi$ iff $(M,s) \not\models \varphi$.

We will often view a Kripke structure $M$ as a directed labeled graph, with
 $S$ as its set of nodes, and whose set of arcs
contains $(s,i,t)$ iff $(s,t) \in \rk_i$. 
$(s,i,t)$ is referred to as an \emph{arc}, from  
the state $s$ to the state $t$. We identify special types
of Kripke structures depending on the properties of $\rk_i$:
\begin{list}{$\bullet$}{\topsep=1pt \parsep=0pt \itemsep=1pt}
\item $M$ is $r$ if, for each $i \in \cala$, 
  $\rk_i$ is reflexive relation;
\item $M$ is $rt$ if, for each $i \in \cala$, 
  $\rk_i$ is reflexive and transitive;
\item $M$ is $rst$ if, for each $i\in \cala$,  
  $\rk_i$ is reflexive, symmetric, and transitive;
\item $M$ is $elt$ if, for each $i\in \cala$, 
  $\rk_i$ is transitive, Euclidean (i.e., for all $s_1,s_2,s_3$,
  if $(s_1,s_2)\in \rk_i$ and $(s_1,s_3)\in \rk_i$ then $(s_2,s_3)\in \rk_i$), and 
  serial (i.e., for each $s$ there exists $s'$ such that 
  $(s,s')\in \rk_i$).
\end{list}
We use $M[S]$, $M[\pi]$, and $M[i]$,
to denote the components $S$, $\pi$, and $\rk_i$ of $M$.
\section{A Simple Action Language for Multi-agent Domains}
\label{sec3}

\iffalse
In recent years, there has been a growth of interest towards
the use of languages like $\lk_{\cala}$ as the foundations for 
modeling actions and interactions among agents. In this section
we illustrate a simple action language whose semantics builds
on a transition system among models of  $\lk_{\cala}$ theories.
\fi

In \cite{BaralGPS10b}, we proposed an action language $\mA$ for multi-agent domains 
that considers various types of actions, such as world-altering actions, announcement actions, 
and sensing actions. In this language, a multi-agent theory
is specified by two components: a $\lk_\cala$ theory and an action 
description. The former describes the initial state of the world and knowledge 
of the agents. The latter describes the actions and their effects. 
We will next briefly describe the syntax of $\mA$. 

%%We need the following 
%%notations. Given a multi-agent domain $\langle \cala, \calf \rangle$, 
%%a fluent literal $\ell$ is either a fluent $f$ or its negation $\neg f$ 
%%for some $f\in \calf$. $f$ and $\neg f$ are said to be {\em complementary} 
%%fluent literals. By $\bar{\ell}$ we denote the complementary fluent literal 
%%of $\ell$. 

\begin{list}{$\bullet$}{\topsep=1pt \parsep=0pt \itemsep=1pt \leftmargin=10pt}

\item The {\em initial state} description is specified by axioms of the form:
\begin{list}{$\circ$}{\topsep=1pt \parsep=0pt \itemsep=1pt \leftmargin=10pt}
%\item \emph{Domain description:} includes statements  of the form
%\[ fluent(f) \hspace{3cm} agent(a) \]
%for each $f\in \calf$ and $a \in \cala$.
\item \emph{Modal logic system:} includes one statement of the form
$system(n)$,
where $n\in \{\mathbf{T}, \mathbf{S4}, \mathbf{S5}, \mathbf{KD45}, \mathbf{none}\}$.

\item \emph{Theory:} includes statements of the form
$init(\varphi)$,
where $\varphi$ is a $\lk_\cala$ formula.
\end{list}
%We will refer to a collection of these statements as a 
%$\lk_\cala$ specification (and we will commonly use $D$ to denote it).

\item The {\em action description} consists of 
%The language mA builds on a multi-agent domain $\langle \cala, \calf\rangle$; 
%a fluent formula is a propositional formula over $\calf$. In particular, a
%fluent literal is of the form $f$ or $\neg f$ with $f\in \calf$. 
%A mA domain specification over the multi-agent domain is a collection of
axioms of the following forms:
\begin{eqnarray} 
& a \executable \delta \label{exec} \\ 
& a \causes \varphi \iif  \psi \performedby \alpha \label{dynamic} \\
& a \announcesk \phi \performedby \alpha \observedby \beta \label{public} \\
& a \determines f  \performedby \alpha \observedby \beta \label{sense} 
\end{eqnarray}
where $a$ is an action, $\alpha, \beta$ are sets of agents from $\cala$, 
$\delta$ is an arbitrary $\lk_\cala$ formula, $\varphi$ and $\psi$ 
are conjunction of fluent literals (a fluent literal is either a fluent $f \in \calf$
or its negation $\neg f$), $\phi$ is a restricted formula (a fluent formula, 
a formula $\k_i\varphi$, or a formula $\neg (\k_i \varphi \vee \k_i(\neg\varphi))$, 
where $\varphi$ is a fluent formula), and $f$ is a fluent.
\end{list}
%If an announcement
%is private, then we restrict $\phi$ to be only a fluent literal.
An axiom of type (\ref{exec}) states the executability condition of the action $a$, 
(\ref{dynamic}) describes a world-altering action, (\ref{public}) an announcement
action, and (\ref{sense}) a sensing action. 
For the sake of simplicity, we assume that the conditions $\psi$ in the
axioms (\ref{dynamic}) for the same action $a$ are mutually exclusive.
An announcement will be referred to as a \emph{public
announcement} if $\alpha = \cala$ and $\beta = \emptyset$, otherwise
it will be referred to as a \emph{private announcement}. 
When it is a private announcement, we restrict $\phi$ to be a fluent literal.
By $(\cala, \calf, D,I)$ we denote a multi-agent theory over  
$\langle \cala, \calf \rangle$ with the initial state $I$ 
and the action description $D$. 

Observe that all axioms describing actions include a part 
indicating the agents participating in the action, 
i.e., who executes the action ({\bf performed\_by}) and who 
is  aware of the action occurrence ({\bf observed\_by}). This is necessary,  
and dealing with this separation is one of the most difficult 
issues in reasoning about knowledge in multi-agent environments \cite{BaltagM04}. 

Given a multi-agent theory $(\cala, \calf, D,I)$, 
we are interested in queries of the form 
\begin{equation} \label{query} 
\varphi \after [a_1;\ldots;a_n] 
\end{equation} 
which asks whether $\varphi$, a $\lk_\cala$ formula,  holds 
after the execution of the action sequence $[a_1,\ldots,a_n]$ from the initial 
state.

\begin{example}\label{example1}
$A$, $B$, and $C$ are in a room. On the table in the middle 
of the room is a box with a lock which contains a coin. 
No one knows whether the head or the tail is up and it 
is common among them that no one knows whether head or tail is up. 
Initially, $A$ and $C$ are looking at the box and $B$ is not. 
One can peek at the coin to know
whether its head or tail is up. To do so, one 
needs a key to the lock and looks at the box. Among the three, 
only $A$ has the key for the lock. 
An agent can make another agent not to look 
at the box by distracting him/her.

{\small\begin{verbatim} 
init(looking_at_box(a)).   init(looking_at_box(c)).  init(~looking_at_box(b)).
init(~k(a,tail)).          init(~k(b,tail)).         init(~k(c,tail)).
init(~k(a,~tail)).         init(~k(b,~tail)).        init(~k(c,~tail)).
init(has_key(a)).          init(~has_key(b)).         init(~has_key(c)).
peek(X,Y)     executable if looking_at_box(X), looking_at_box(Y), has_key(X).
distract(X,Y) executable if true.
peek(X,Y) determines tail performed_by  X observed_by Y.
distract(X,Y) causes ~looking_at_box(Y) if true performed by X.
\end{verbatim}}
\end{example}

\section{Computing the Initial State} 
\label{sec4}

Computing an initial state of a multi-agent theory $(\cala, \calf, D,I)$ means 
to compute a pointed Kripke structure $(M,s)$ 
satisfying $I$ where $(M,s)$ satisfies $I$ if 
\begin{list}{$\bullet$}{\topsep=1pt \parsep=0pt \itemsep=1pt}
\item $(M,s)$ is a pointed Kripke structure w.r.t. the multi-agent domain 
$\langle \cala, \calf \rangle$; 
\item For each $\textit{init}(\varphi)\in I$ we have that
  $(M,s) \models \varphi$;
\item If $\textit{system}(\mathbf{T})\in I$ (resp. $\textit{system}(\mathbf{S4})$, $\textit{system}(\mathbf{S5})$,
  $\textit{system}(\mathbf{KD45})$), then $M$ is $r$ (resp. $rt$, $rst$, $elt$).
\end{list}

\subsection{Computing Initial State Using Answer Set Programming}

In \cite{BaralGSP10a}, we proposed an answer set programming
(ASP) \cite{MarekT99,Niemela99} implementation for computing 
the initial state of multi-agent theories. The implementation follows the 
basic idea of ASP by converting the initial state 
specification $I$ to an ASP program $\Pi_I(m)$, with $m$ being the number of 
states of the structure, whose answer sets are pointed 
Kripke structures satisfying $I$.  The language of $\Pi_I(m)$ includes:
\begin{list}{$\circ$}{\topsep=1pt \parsep=0pt \itemsep=1pt \leftmargin=10pt}
\item A set of facts $\textit{fluent}(f)$ ($\textit{agent}(a)$) for each $f \in \calf$
  (resp. $a \in \cala$);
\item A set of constants $s_1,\ldots,s_n$, representing the names of the possible states;
\item Atoms of the form $state(S)$, denoting the fact that $S$ is a state in the
  Kripke structure being built;
\item Atoms of the form $h(\varphi,S)$, indicating  that the formula 
  $\varphi$ holds in the state $S$ in the Kripke   structure.
 These atoms represent the interpretation $\pi$ associated with each state  and 
  the set of formulae entailed by the pointed Kripke structure;
\item Atoms of the form $r(A,S_1,S_2)$, representing the accessibility 
relations of the Kripke structure, i.e., $(S_1,S_2) \in \rk_A$ in the Kripke structure;
\item Atoms of the form $t(S_1, S_2)$, used to represent the existence of 
a path from $S_1$ to $S_2$ in the  Kripke  structure present;
\item Atoms of the form $real(S)$, to denote the real state of
 the world.
\end{list}
We assume formulae to be built from fluents, propositional connectives, 
and modal operators (in keeping with our previous definition). 
In particular, the fact that a formula of the 
form $\k_A\varphi$ holds in the current Kripke structure with respect to a state $S$ is encoded 
by atoms of the form $k(A,  S, \varphi)$.  

The creation of the model requires selecting which states and which connections should be
included in the Kripke structure and is expressed by choice rules:
\[\begin{array}{lll}
  1 \{ state(s_1),\dots,state(s_n)\}m  & \leftarrow & \\
  0 \{h(F,S)\} 1 & \leftarrow &   fluent(F), state(S) \\
  0 \{ r(A,S_1, S_2) : state(S_1) : state(S_2) \} 1 & \leftarrow &  \textit{agent}(A)\\
  1 \{ real(S) : state(S) \} 1 & \leftarrow & 
\end{array}\]
These rules will generate a pointed Kripke structure $(M,s)$ whose states belong to 
$\{s_1,\ldots,s_n\}$ with $s$ indicated by $real(s)$. 
The program is completed with rules determining the truth 
value of a formula in a given pointed structure. For instance, 
rules for checking whether an agent $A$ knows a literal $L$ are:\footnote{
    \cite{BaralGSP10a} contains rules for checking whether $(M,s)$ entails other types of formulae. 
}  
\[\begin{array}{lll}
n\_k(A, S, F) & \leftarrow &  r(A, S, S_1), h(\neg F, S_1) \\
n\_k(A, S, \neg F) & \leftarrow & r(A, S, S_1), h(F, S_1) \\
k(A, S, L) &  \leftarrow & \naf \:\:\:\:n\_k(A, S, L)
\end{array}
\]
In addition to the above rules, $\Pi_I(m)$ also contain rules encoding 
the additional properties imposed by a specific modal logic system: 
\[
\begin{array}{rcl} 
r(A, S,S) & \leftarrow & reflexivity, state(S), agent(A) \\
r(A,S_1, S_2) & \leftarrow & symmetry, r(A, S_2, S_1) \\
r(A,S_1,S_3) & \leftarrow & transitivity, r(A,S_1,S_2), r(A,S_2,S_3) \\
1 \{r(A,S,T): state(T)\} & \leftarrow & serial, state(S) \\
r(A,S_2,S_3) & \leftarrow & euclidean, r(A,S_1,S_2), r(A,S_1,S_3) 
\end{array} 
\] 

\iffalse 
\begin{itemize} \itemsep=0pt \leftmargin=10pt
\item If reflexivity is required, then we add
\[ r(A, S,S) \leftarrow reflexivity, state(S), agent(A)\]
\item If symmetry is required, we add
\[ r(A,S_1, S_2) \leftarrow symmetry, r(A, S_2, S_1)\]
\item If transitivity is required, we add
\[ r(A,S_1,S_3) \leftarrow transitivity, r(A,S_1,S_2), r(A,S_2,S_3)\]
\item If seriality is required, we add
\[ 1 \{r(A,S,T): state(T)\} \leftarrow serial, state(S)\]
\item If Eucledian behavior is required, add
\[ r(A,S_2,S_3) \leftarrow euclidean, r(A,S_1,S_2), r(A,S_1,S_3)\]
\end{itemize}
\fi 

We can activate the necessary rules according to the modal logic system; e.g.,
if $\textit{system}(\mathbf{S4})\in I$, then we add the facts $reflexive$ and $transitive$ to 
$\Pi_I(m)$.

Finally, in order to ensure that a pointed structure satisfying $I$ is generated, we add to $\Pi_I(m)$ 
the constraint: $\leftarrow init(\varphi), real(S), not\:h(\varphi,S).$

The program $\Pi_I(m)$ can be used not only to generate models of $I$ but also 
to retrospectively identify properties of $I$ given a history of action 
occurrences. In \cite{BaralGSP10a}, we used $\Pi_I(m)$ to solve the muddy children 
problem.

For example, one of the Kripke structures generated for the initial specification of 
Example~\ref{example1} includes: \emph{(1)} two states ($S=\{s_1,s_2\}$), 
\emph{(2)} the
interpretations $\pi(s_1)$ and $\pi(s_2)$ make {\tt looking\_at\_box}
true for $a,c$ and false, {\tt has\_key} true for $a$ and false for $b,c$,
and they differ on the truth value of {\tt tail} (e.g.,
{\tt tail} is true in $\pi(s_1)$ and false in $\pi(s_2)$), and
\emph{(3)} $(s_i,s_j)\in \rk_x$ for each $x\in \{a,b,c\}$ and
$i,j \in \{1,2\}$.

\subsection{Computing Initial State Using Prolog}
The elegance of ASP encoding has unfortunately to deal with 
the complexity of grounding imposed by modern ASP systems---and this
will motivate our contribution of using Prolog for this task. 
Indeed, it is
possible to find even simple theories of logics of knowledge that are
beyond the capabilities of ASP. 
%This can be seen in the following example:

%Let us consider an extension of our language by allowing addition
%of announcements---i.e., addition of new formulae
%that have to be satisfied, written as
%\emph{announce($\varphi$)}. 

\begin{example}[Sum and Product]
An agent chooses
two numbers $1 < x < y$ such that $x+y \leq 100$. The sum $x+y$
is communicated to agent $s$ while the product $x*y$ is communicated
to agent $p$. The
following conversation takes place between the two agents:
\begin{list}{$\bullet$}{\topsep=1pt \parsep=0pt \itemsep=1pt \leftmargin=12pt}
\item  $p$ states that it does not know the numbers $x$, $y$
\item $s$ indicates that it already knew this fact
\item  $p$ states that now it knows the two numbers $x$, $y$
\item  $s$ states that now it knows the two numbers $x$, $y$ as
well.
\end{list}
\end{example}
Let us consider the problem of generating an initial Kripke structure 
for the above story. The two agents are operating on states 
that can be represented by a pair $(x,y)$ satisfying the given conditions. 
It is reasonable to assume that initially, the agents will need to 
consider {\em all} possible states. 
This is encoded in ASP by the rule:
\[
state(X,Y) \leftarrow 1 < X, X < Y, X+Y < 101
\]
It is easy to see that the number of states is 2352. To generate 
the initial Kripke structure using ASP, we would have to use the rule 
\[
0 \{r(A,state(X,Y),state(X_1,Y_1),0) : state(X,Y) : state(X_1,Y_1) \} 1 \leftarrow agent(A) 
\]
which will produce $2352^2$ ground rules for each agent during the grounding. 
Computing a model for a program consisting of only this rule and the rule
defining the states is already impossible.\footnote{ 
   We got the ``Out of memory'' message in Clingo. Lparse did not finish. 
} 
A reasonable (but ad-hoc) way is to use the rules 
\[\begin{array}{l}
r(s, state(X,Y), state(X_1,Y_1), 0) \leftarrow X+Y=X_1+Y_1, state(X,Y), state(X_1,Y_1)\\
r(p, state(X,Y), state(X_1,Y_1), 0) \leftarrow X*Y=X_1*Y_1, state(X,Y), state(X_1,Y_1)\\
\end{array}
\]
which reduces the number of ground rules to less than $2352 \times 110$ 
($2352 \times 99$ for the sum, $2352 \times 10$ for the product) in total. Intuitively, these 
rules indicate that there is a link labeled $s$ between $state(x,y)$ and $state(x',y')$ 
iff $x+y = x'+y'$, and  
 there is a link labeled $p$ between $state(x,y)$ and $state(x',y')$ iff $x\times y = x' \times y'$.
Using these rules, Clingo/Smodels\footnote{\url{potassco.sourceforge.net}, \url{www.tcs.hut.fi/Software/smodels}} is able to find a model (i.e., a possible Kripke structure)
within a few seconds. 

In order to verify that the generated Kripke structure satisfies the initial statements 
of the story, we will need to introduce fluents of the form $sum(S)$
and $product(S)$ to denote the sum and product of the two numbers of a state
$S=(X,Y)$. We would also need to have fluents of the form $x(S)$ and $y(S)$ 
to represent that $x$ and $y$ are the two components of the state $S$.

The rules describing the truth values of the fluents are: 
\[ \begin{array}{rlr}
h(x(X), state(X,Y))  \leftarrow  & & h(sum(S), state(X,Y))  \leftarrow S = X+Y \\ 
h(y(Y), state(X,Y)) \leftarrow   & & h(product(P), state(X,Y))  \leftarrow P = X*Y\\
\end{array}
\]
%Observe that the rules are presented with a parameter $T$, indicating the 
%time step in the evolution of the story. 
%The values of the fluents remain 
%unchanged due to the fact that there are no actions that change the value of fluents.  
With these fluents, we can define a formula stating that an agent knows
the value of the two numbers by 
$ s\_knows \equiv \bigvee_{state(x,y)} K_s(x(X)\wedge y(Y))$
and 
$ p\_knows \equiv \bigvee_{state(x,y)} K_p(x(X)\wedge y(Y))$.

The initial pointed Kripke structure is one that satisfies the 
formula $\neg p\_knows$ and $k_s(\neg p\_knows)$. This is what $p$ and $s$ state in 
their first announcement, respectively. Thus, to construct the initial pointed Kripke structure, we will 
need to add the rules defining the predicate $k(Agent, State, Formula)$ 
and other rules relating to this predicate. We will also need to add the constraints   
\[
\begin{array}{ll} 
\leftarrow real(X,Y), \naf \:\:k(p, state(X,Y), p\_knows) \\
\leftarrow real(X,Y), \naf\:\: k(s, state(X,Y), \neg k(p, state(X,Y), p\_knows)) 
\end{array} 
\]
The first constraint corresponds to $p$'s first statement and the second to $s$'s first statement. 
Adding these rules to the program, the ASP program does not find a model within two hours.
The main culprit is the number of ground rules that need to be generated 
before the answer set can be computed. For instance, the number of rules defining $n\_k$, 
(as described in the previous section) is roughly $2352^2$ (quadratic 
to the number of the states). 

Many of these complications can be avoided by changing the
model of computation from a bottom-up model (as used by ASP) to
a top-down one (as used by Prolog). 
The Prolog encoding builds the similar clauses as described for
ASP, to verify validity of a formula in a Kripke structure. The
advantage is that, by operating top-down, the components of the
structure are computed when requested (instead of being precomputed
a priori during grounding).

We will now present a Prolog encoding for computing the initial pointed Kripke structure 
for a multi-agent theory. Each Kripke structure can be encoded using terms of the form
{\tt kripke(N,E)} where {\tt N} is a list of nodes and {\tt E} a list of
edges. Each node is a term {\tt node(Name, Interpretation)}, where {\tt Name}
is a name for the node and the interpretation is encoded as a list
of terms {\tt value(Fluent,true/false)}. A pointed Kripke structure is
encoded as a term {\tt sit(kripke(N,E), Node)} where {\tt Node}
is an element from the list {\tt N}. The clauses to express the validity
of a formula are very similar to the one presented earlier, e.g.,\footnote{The
explicit representation of interpretations can be easily replaced with 
implicit encodings whenever specific domain knowledge is available.}

{\small\begin{verbatim}
hold(F,sit(kripke(_Nodes,_Edges),N)) :-     
  fluent(F),!, N = node(_Name,Int), member(value(F,true), Int). 
hold(neg(F), Sit) :- \+hold(F,Sit), !.
hold(k(A,F), sit(kripke(N,E),node(X,_))) :-     
  findall(M,member(edge(X,M,A),E),Nodes), iterated_hold(Nodes,F,kripke(N,E)). 
iterated_hold([],_,_). 
iterated_hold([A|B],F, kripke(N,E)) :-     
  member(node(A,I),N), hold(F, sit(kripke(N,E),node(A,I))),     
  iterated_hold(B,F,kripke(N,E)). 
\end{verbatim}}

In the sum-and-product example, the initial Kripke structure can
be implicitly defined using implicit rules to construct the edges of
the graph:

{\footnotesize\begin{verbatim}
edge(node(N1,Int1),node(N2,Int2),s) :- 
   member(value(x(X),true),Int1), member(value(y(Y),true),Int1),
   member(value(x(X1),true),Int2), member(value(y(Y1),true), Int2), X+Y =:= X1+Y1.
edge(node(N1,Int1),node(N2,Int2),p) :- 
   member(value(x(X),true),Int1), member(value(y(Y),true),Int1),
   member(value(x(X1),true),Int2), member(value(y(Y1),true), Int2), X*Y =:= X1*Y1.
\end{verbatim}}

The Prolog implementation of the announcements as constraints on the 
Kripke structure allows us to quickly converge
on identifying $x=4$ and $y=13$ as the real state of the pointed Kripke
structure in a matter of seconds ($10.5$ seconds on a MacBook Pro, 2.53GHz core duo).

\section{The Semantics of $\mA$}
\label{sec5}

We will now describe the semantics of $\mA$, based on the
construction of a transition function. For the sake of simplicity,
in this manuscript we will restrict our discussion to domains where
the initial state of the domain is described by one pointed Kripke
structure and the actions are deterministic. Thus, the transition
function is a map from a pointed Kripke structure and an action
to another pointed Kripke structure.\footnote{It is easy to generalize
this to maps where the possible configurations are described by 
sets of pointed Kripke structures and the actions are 
non-deterministic.}

\subsection{Basic Kripke Structure Transformations}

We start by introducing some basic transformations of 
Kripke structures necessary to model the evolution in models caused
by the execution of actions. We also show how, following the representation of Kripke
structures discussed in Section~\ref{sec2}, it is possible to naturally encode
these operators in Prolog.

Given a Kripke structure $M$, a set of states $U \subseteq M[S]$, and
a set of arcs $X$ in $M$,
  $M \stateminus U$ is the Kripke structure $M^{\prime}$ defined by
({\em i}) $M^{\prime}[S] = M[S] \setminus U$;
({\em ii}) $M^{\prime}[\pi](s)(f) = M[\pi](s)(f)$ for every $s \in M^{\prime}[S]$
	and $f \in \calf$;
and ({\em iii}) $M^{\prime}[i] = M[i] \setminus \{ (t,v) \mid (t,v) \in
M[i], \{t,v\} \intersect U \neq \emptyset \}$ for every agent $i \in \agents$.
Intuitively, $M \stateminus U$ is the Kripke structure obtained by removing from
$M$ all the states in $U$.

The $\stateminus$ operator is encoded by the following Prolog rules:

{\small\begin{verbatim}
node_minus(kripke(N,E), S, kripke(N1,E1)) :-
        delete(N,S,N1), remove_node_edges(S,E,E1).
remove_node_edges([],E,E).
remove_node_edges([A|B], E, NewEs) :-
        remove_one_node_edges(A,E, Es1), remove_node_edges(B, Es1, NewEs).
remove_one_node_edges(_,[],[]).
remove_one_node_edges(node(Node,Int), [edge(N1,N2,_)|Rest], Result) :-
        (N1=Node ; N2 = Node), !,
        remove_one_node_edges(node(Node,Int), Rest, Result).
remove_one_node_edges(N,[E|Rest],[E|Result]) :-
        remove_one_node_edges(N,Rest,Result).
\end{verbatim}}

$M \arcminus X$ is a Kripke structure, $M^{\prime}$, defined by
({\em i}) $M^{\prime}[S] = M[S]$; ({\em ii}) $M^{\prime}[\pi] = M[\pi]$;
and ({\em iii}) $M^{\prime}[i] = M[i] \setminus \{ (u,v) \mid (u,i,v) \in X \}$ for
 $i \in \agents$.
$M \arcminus X$ is the Kripke structure obtained by removing
from $M$ all the arcs in $X$. 
The encoding of $\arcminus$ is:

{\small\begin{verbatim}
     edge_minus(kripke(N,E), Es, kripke(N,E1)) :- delete(E,Es,E1).
\end{verbatim}}

Given two Kripke structures $M_{1}$ and $M_{2}$, we say that
$M_{1}$ is \textit{c-equivalent}\footnote{
   This form of equivalence is similar to the notion of bisimulation in \cite{BaltagM04}.
} to $M_{2}$\ ($M_{1} \cequiv M_{2}$) if
there exists a bijective function $c : M_{1}[S] \rightarrow M_{2}[S]$ such that:
\emph{(i)} for every $u \in S$ and $f \in \fluents$,  $M_{1}[\pi](u)(f) = true$
iff $M_{2}[\pi](c(u))(f) = true$; and \emph{(ii)}
for every $i \in \agents$ and $u,v \in M_{1}[S]$, $(u,v) \in M[i]$
iff $(c(u),c(v)) \in M_{2}[i]$.

$M_{1}$ and $M_{2}$ are \textit{compatible}
if for every $s \in M_{1}[S] \intersect M_{2}[S]$ and every
$f \in \calf$, $M_{1}[\pi](s)(f) = M_{2}[\pi](s)(f)$.

For two compatible Kripke structures $M_1$ and $M_2$, we define
$M_{1} \ckplus M_{2}$ to be the Kripke structure $M^{\prime}$, where
({\em i}) $M^{\prime}[S] = M_{1}[S] \union M_{2}[S]$;
({\em ii}) $M^{\prime}[\pi] = M_{1}[\pi] \cat M_{2}[\pi]$;\footnote{More precisely,
$M'[\pi](s)(f) = M_1[\pi](s)(f)$ if $s \in M_1[S]$ and
$M'[\pi](s)(f) = M_2[\pi](s)(f)$ if $s \in M_2[S]\setminus M_1[S]$.
   } and ({\em iii}) $M^{\prime}[i] = M_{1}[i] \union M_{2}[i]$.
   The encoding in Prolog of $M_{1} \ckplus M_{2}$ is

{\small\begin{verbatim}
union_kripke(kripke(N1,E1), kripke(N2,E2), kripke(N3,E3)) :-
    append(N1,N2,N4), remove_dups(N4,N3), append(E1,E2,E4), remove_dups(E4,E3).
\end{verbatim}   }

For a pair of Kripke structures $M_{1}$ and $M_{2}$ with
$\disjoint{M_{1}[S]}{M_{2}[S]}$, $\alpha \subseteq \agents$,
and a one-to-one function $\lambda : M_{2}[S] \rightarrow M_{1}[S]$,
we define $M_{1} \kplus{\lambda}{\alpha} M_{2}$
to be the Kripke structure $M^{\prime}$, where
({\em i}) $M^{\prime}[S] = M_{1}[S] \union M_{2}[S]$;
({\em ii}) $M^{\prime}[\pi] = M_{1}[\pi] \cat M_{2}[\pi]$;
({\em iii})  $M^{\prime}[{i}] = M_{1}[{i}] \union M_{2}[{i}]$ for each
${i} \in \alpha$, and $M^{\prime}[i] = M_{1}[i] \union M_{2}[i] \union
\{ (u,v) \mid u \in M_{2}[S], v \in M_{1}[S], (\lambda(u),v) \in
M_{1}[i]\}$ for each agent
$i \in \agents \setminus \alpha$.

Intuitively, the operators $\ckplus$ and $\kplus{\lambda}{\alpha}$ allow
us to combine different Kripke structures representing knowledge
of different groups of agents, thereby creating a structure representing
the knowledge of all the agents. The encoding of $\kplus{\lambda}{\alpha}$ in Prolog is:

{\small\begin{verbatim}
knowledge_union(kripke(N1,E1), kripke(N2,E2), Alpha, Map, kripke(N3,E3)) :-
    check_k_union_properties(N1,N2,Map), union_list(N1,N2,N3),
    generate_k_union_edges(E1, N2, Alpha,Map, NewOnes),
    union_list(E1,E2,E4), union_list(E4,NewOnes,E3).
generate_k_union_edges([], _, _, _, []).
generate_k_union_edges([edge(_,_,Label)|Rest],Nodes,Alpha,Map,NRest) :-
    member(Label,Alpha), !, generate_k_union_edges(Rest,Nodes,Alpha,Map,NRest).
generate_k_union_edges([edge(Start,End,Label)|Rest], Nodes, Alpha, 
                                                 Map, [NEdge|NRest]) :-
    member([N,Start], Map), NEdge = edge(N,End,Label),
    generate_k_union_edges(Rest,Nodes,Alpha,Map,NRest).
\end{verbatim}}

Several types of actions require the creation of
``copies'' of a  Kripke structure, typically to encode
the knowledge of the agents that are unaware of actions being executed.
Let
$\situation{M}{s}$ be a pointed Kripke structure, and $\alpha$ be a set of agents:
\begin{list}{$\bullet$}{\topsep=1pt \parsep=0pt \itemsep=1pt}
    \item A pointed Kripke structure $\situation{M^{\prime}}{c(s)}$ is a
\textit{replica} of $\situation{M}{s}$ if $M^{\prime} \cequiv M$ and
$\disjoint{M^{\prime}[S]}{M[S]}$;
    \item $\situation{M}{s}|_{\alpha} = \situation{M \arcminus X}{s}$
where $X = \Union_{{i} \in \alpha}\{ (u,{i},v) \mid
(u,v) \in M[{i}] \}$;
\end{list}
A replica of a pointed Kripke structure $\situation{M}{s}$ refers to a
copy of $\situation{M}{s}$ with respect to a bijection $c$. The
pointed Kripke structure $\situation{M}{s}|_{\alpha}$, referred to as \textit{the
restriction of $\situation{M}{s}$ on $\alpha$}, encodes the knowledge
of the agents in the set $\agents \setminus \alpha$. In Prolog:

{\small\begin{verbatim}
replica(kripke(N,E), kripke(N1,E1), Map) :-
    replica_nodes(N,N1,Map), replica_edges(E,Map,E1).
replica_nodes([],[],[]).
replica_nodes([node(Name,Int)|Rest], [node(NewName,Int)|NewRest], 
                     [[NewName,Name]|RestMap]) :-
    new_state_name(NewName), replica_nodes(Rest,NewRest,RestMap).
replica_edges([],_,[]).
replica_edges([edge(Start,End,Agent)|Rest], Map, 
                     [edge(NewStart,NewEnd, Agent)|NewRest]) :-
    member([NewStart,Start], Map), member([NewEnd,End],Map),
    replica_edges(Rest,Map,NewRest).
remove_agent_edges(kripke(N,E), Alpha, K) :-
    collect_agent_edges(E,Alpha,Edges), edge_minus(kripke(N,E), Edges, K).
collect_agent_edges([],_,[]).
collect_agent_edges([edge(Start,End,Agent)|Rest], Alpha, 
                            [edge(Start,End,Agent)|NewRest]) :-
    member(Agent,Alpha),!, collect_agent_edges(Rest,Alpha,NewRest).
collect_agent_edges([_|Rest], Alpha, NewRest) :-
    collect_agent_edges(Rest,Alpha,NewRest).
\end{verbatim}}

\subsection{Action Transition Function}

The following definitions are used to determine, given a pointed Kripke structure $(M,s)$,
what are the pointed Kripke structures obtained from the execution of an action.
We will denote with $succ(a, (M,s))$ the pointed Kripke structure
resulting from the execution of the action $a$
 in $(M,s)$. Observe
that in the following definitions we view a set of literals as the conjunction of its
elements. We will define $succ(a, (M,s))$ for each 
type of action. 
Let $a$ be an action. By $pre(a)$ we denote the formula $\delta$, 
the condition in the law of the from (\ref{exec}) whose action is $a$. 
We begin with the public announcement action. 

\medskip \noindent 
{\em Public Announcement:} Consider a pointed Kripke structure $\situation{M}{s}$ and an action
instance  $a$
occurring in an announcement law (\ref{public}) such that  $\situation{M}{s} \models \phi$.
The successor pointed Kripke structure after the execution of $a$ 
in $\situation{M}{s}$ is defined as follows.
If $(M,s) \not\models pre(a)$, then $\successor{a}{\situation{M}{s}}$ is
undefined, denoted by $\successor{a}{\situation{M}{s}} = \bottom$. The first case deals with 
the assumption that the action should be executable 
in
$\situation{M}{s}$---otherwise 
the resulting successor pointed Kripke structure is undefined. 

If $\situation{M}{s} \entails \phi$ then we have different cases depending on the
structure of $\phi$.
If $\phi$ is a fluent formula then $\successor{a}{\situation{M}{s}} =
\situation{M \stateminus U}{s}$ for 
$U = \{ s^{\prime} \mid s^{\prime} \in M[S], \situation{M}{s^{\prime}} \not\entails \phi
\}.$ This can be captured by the following rules:

{\small\begin{verbatim}
succ(sit(kripke(N,E),S), public(Phi), sit(kripke(N1,E1),S)) :-
    fluent_formula(Phi), !,
    get_nodes_not_satisfy_formula(Phi, kripke(N,E), Nodes),
    node_minus(kripke(N,E), Nodes, kripke(N1,E1)).
get_nodes_not_satisfy_formula(Phi, kripke(N,E), Nodes) :-
    iterate_not_satisfy(N,Phi,kripke(N,E), Nodes).
iterate_not_satisfy([],_,_,[]).
iterate_not_satisfy([A|B],F,K,Rest) :-
    hold(F,sit(K,A)), !, iterate_not_satisfy(B,F,K,Rest).
iterate_not_satisfy([A|B], F, K, [A|Rest]) :- iterate_not_satisfy(B,F,K,Rest).
\end{verbatim}}

If $\phi = \k_{i}\psi$, then $\successor{a}{\situation{M}{s}} = \situation{M
\arcminus X}{s}$ where $X = \{ (u,i,v) \mid (u,v) \in M[i],
\situation{M}{v} \not\entails \psi \}$. This is captured by the following
rules:

{\small\begin{verbatim}
succ(sit(kripke(N,E),S), public(Phi), sit(kripke(N1,E1),S)) :-
    Phi = k(A,Psi), fluent_formula(Psi), !,
    get_edges_not_satisfy(E, A, Psi, kripke(N,E), Edges),
    edge_minus(kripke(N,E), Edges, kripke(N1,E1)).
get_edges_not_satisfy([],_,_,_, []).
get_edges_not_satisfy([Edge|Rest], Agent, Psi, kripke(N,E),[Edge|Result]) :-
    Edge=edge(Start,End,Agent), member(node(End,Int), N),
    \+ hold(Psi, sit(kripke(N,E), node(End,Int))), !,
    get_edges_not_satisfy(Rest,Agent,Psi,kripke(N,E), Result).
get_edges_not_satisfy([_|Rest],Agent,Psi,K,Result) :-
    get_edges_not_satisfy(Rest,Agent,Psi,K,Result).
\end{verbatim}}

If $\phi = \neg(\k_{i}\psi {\disj} \k_{i}\neg{\psi})$, then
$\successor{a}{\situation{M}{s}} = \situation{M \stateminus U}{s}$
where $U {=} \{ s^{\prime} \mid s^{\prime} \in M[S],
\situation{M}{s^{\prime}} \entails \k_{i}\psi \disj
\k_{i}\neg{\psi}\}$. We omit the Prolog code due to lack of space.

If the announcement happens in
a pointed Kripke structure $\situation{M}{s}$ where $\situation{M}{s} \entails
\varphi$, then we need to ensure that $\varphi$ is common knowledge (strictly speaking, 
this is the belief of the agents) among
the agents of the system. This accounts for our removal of all nodes
$s^{\prime}$ such that $\situation{M}{s^{\prime}} \not\entails
\varphi$. Similarly, if a formula $\k_{i}\psi$ is announced, we
only remove arcs labeled by $i$ from $M$ whose existence
invalidates the formula $\ck_{\agents} \k_{i}\psi$. On the other
hand, when a formula $\neg(\k_{i}\psi \disj \k_{i}\neg{\psi})$
is announced, we need to remove states in $M$ whose existence
invalidates the formula $\ck_{\agents}\neg( \k_{i}\psi \disj
\k_{i}\neg{\psi})$.

\begin{figwindow}[0,r,%
\fbox{\epsfig{figure=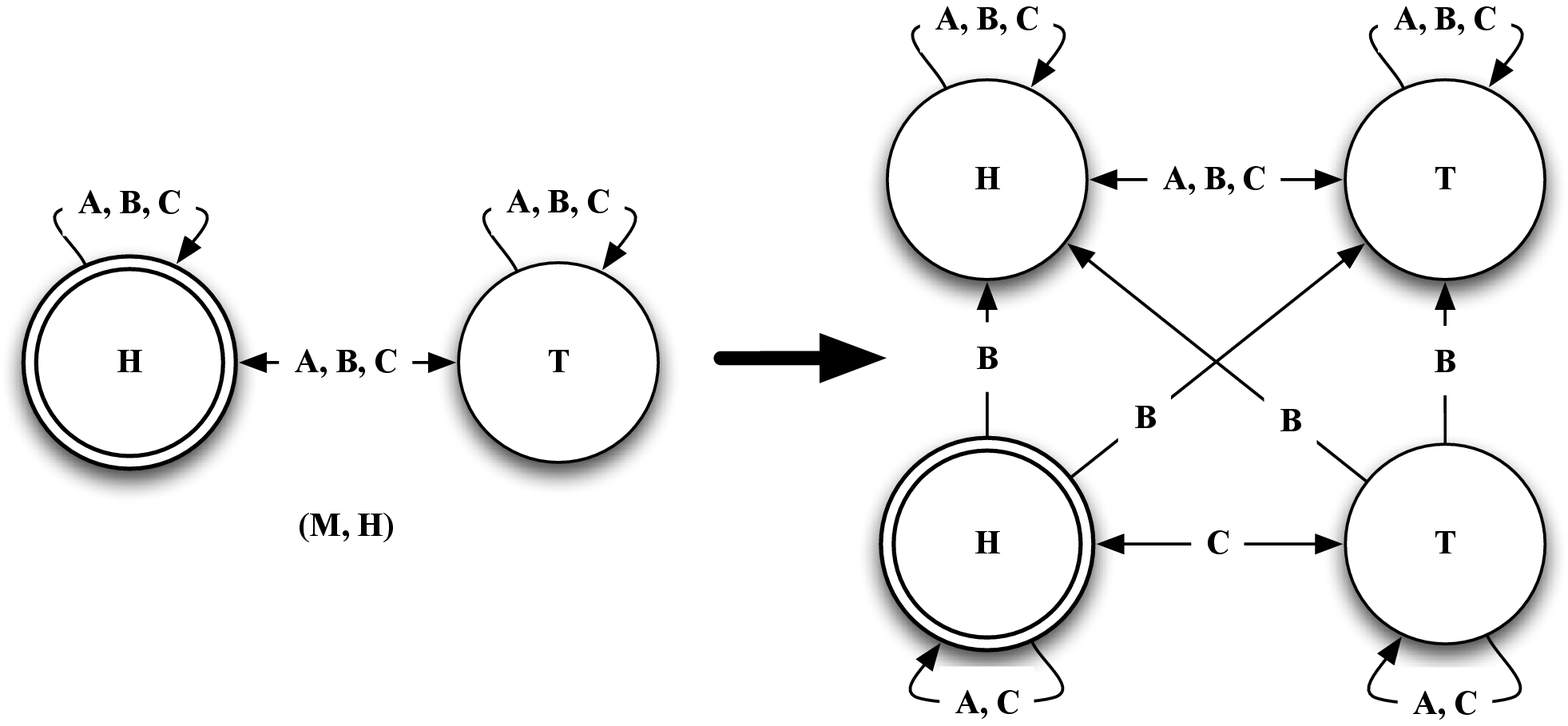,width=.4\textwidth}},%
{Execution of  {\tt peek(a,c)}\label{trans}}]
Continuing with Example~\ref{example1},
the execution of the action {\tt peek($a$,$c$)} in the initial
Kripke structure described earlier transforms the
Kripke structure as illustrated in Figure~\ref{trans},  where the double circle
identifies the real state of the world, $H$ denotes a state
where {\tt tail} is false and $T$ denotes a state where 
{\tt tail} is true. 
\end{figwindow}

\medskip \noindent 
{\em Private Announcement:} Consider a pointed Kripke structure $\situation{M}{s}$ and a
private announcement action $a$ with 
$\beta \neq \cala$, where  the effect is the literal $\ell$ and
$\situation{M}{s}\models \ell$. 
Let us also assume that $f$ is
the fluent used in the literal $\ell$.
The pointed Kripke structure
$\situation{M^{\prime}}{s^{\prime}}$ is the successor model after
the execution of $a$  in $\situation{M}{s}$ if:
\begin{list}{$\bullet$}{\topsep=1pt \parsep=0pt \itemsep=1pt}
    \item $(M,s) \models pre(a)$ then 
    $
    \situation{M^{\prime}}{s^{\prime}} = \situation{M}{s}
\uplus^{c}_{\beta \union \gamma}(M^{r}|_{\agents \setminus (\beta
\union \gamma)} \arcminus X, c(s))
    $
    where $\situation{M^{r}}{c(s)}$ is a replica of $\situation{M}{s}$
and
$X = \left\{ (u,{i},v) \begin{array}{|l}
                {i} \in \beta, (u,v) \in M[{i}], \\
                M[\pi](u)(f) \neq M[\pi](v)(f)
                \end{array}\right\}$;
    \item $\situation{M}{s} \not\entails pre(a)$ and
$\situation{M^{\prime}}{s^{\prime}} = \bottom$.
\end{list}
The general successful case can be captured as:
{\small\begin{verbatim}
succ(sit(kripke(N,E),S), private(Phi,Alpha,Beta), sit(kripke(N1,E1),St1)) :-
    get_literal_fluent(Phi,F), replica(kripke(N,E), kripke(N2,E2), Map), 
    non_agents(Alpha,Beta,RemAgs), 
    remove_agent_edges(kripke(N2,E2), RemAgs, kripke(N3,E3)),
    catch_discriminating_edges(E3,kripke(N3,E3),Alpha,F, X),
    edge_minus(kripke(N3,E3), X, kripke(N4,E4)),
    append(Alpha,Beta,AlphaBeta), 
    knowledge_union(kripke(N,E), kripke(N4,E4), AlphaBeta, Map,kripke(N1,E1)),
    S = node(NameS,_), member([S1, NameS],Map), member(node(S1,Int1),N1),
    St1 = node(S1,Int1).
\end{verbatim}}

The predicate {\tt non\_agents} collects all agents not in 
{\tt Alpha} and {\tt Beta}. The predicate {\tt catch\_discriminating\_edges}
identifies edges for which the agents in {\tt Alpha} see distinct truth
values for the considered literal:

{\small\begin{verbatim}
catch_discriminating_edges([],_,_,_,[]).
catch_discriminating_edges([Ed|Rest], kripke(N,E), Alpha, F, [Ed|NewRest]) :-
    Ed = edge(Start,End,Agent), member(Agent,Alpha), 
    member(node(Start,Int1), N), member(node(End,Int2), N),
    member(value(F,V1), Int1), member(value(F,V2), Int2), V1 \= V2, !,
    catch_discriminating_edges(Rest, kripke(N,E), Alpha, F, NewRest).
catch_discriminating_edges([_|Rest], K, Alpha, F, NewRest) :-
    catch_discriminating_edges(Rest,K,Alpha,F,NewRest).
\end{verbatim}}

\medskip \noindent 
{\em Sensing:}
Let $\situation{M}{s}$ be a pointed Kripke structure and $a$ be a
sense action. A model
$\situation{M^{\prime}}{s^{\prime}}$ is a successor model after
the execution of $a$ in $\situation{M}{s}$ if:\\
\centerline{$
    \situation{M^{\prime}}{s^{\prime}} = \situation{M}{s}
\uplus^{c}_{\alpha \union \beta}(M^{r}|_{\agents \setminus (\alpha
\union \beta)} \arcminus X, c(s))
$}
for $X = \{ (u,{i},v) \mid {i} \in \alpha, (u,v) \in
M^{r}[{i}], M^{r}[\pi](u)(f) \neq M^{r}[\pi](v)(f) \}$ and
some replica $\situation{M^{r}}{c(s)}$ of $\situation{M}{s}$.
The Prolog code is similar to the private
announcement:

{\small\begin{verbatim}
succ(sit(kripke(N,E),S), sense(F,Alpha,Beta), sit(kripke(N1,E1),St1)) :-
    replica(kripke(N,E), kripke(N2,E2), Map), non_agents(Alpha,Beta,RemAgs),
    remove_agent_edges(kripke(N2,E2), RemAgs, kripke(N3,E3)),
    catch_discriminating_edges(E3,kripke(N3,E3),Alpha,F, X),
    edge_minus(kripke(N3,E3), X, kripke(N4,E4)), append(Alpha,Beta,AlphaBeta), 
    knowledge_union(kripke(N,E), kripke(N4,E4), AlphaBeta, Map,kripke(N1,E1)),
    S = node(NameS,_), member([S1, NameS],Map),
    member(node(S1,Int1),N1), St1 = node(S1,Int1).
\end{verbatim}}

\medskip \noindent 
{\em World-altering Action:} the computation of the successor pointed
Kripke structure for world-altering actions requires the modification
of the interpretations associated to certain states. 
Given an interpretation $\pi$ and a set of literals $\varphi$, let us denote
with $[\varphi]\pi$ the interpretation obtained by performing the minimal amount
of modifications to $\pi$ to ensure that $[\varphi]\pi$ makes all the literals
in $\varphi$ true.
Let $u\in M[S]$  and let us consider an axiom of type (\ref{dynamic}); the axiom
is applicable in $u$  if $(M,u)\models \psi$.
We  define
\[
res(a, u) =
\left \{
\begin{array}{lll}
[\varphi]M[\pi](u)  &
  \textnormal{if $a \:\causes \: \varphi \:\iif \: \psi\: \performedby\: \alpha$ applicable in $u$}\\
M[\pi](u) &\textnormal{otherwise}
\end{array}
\right.
\]
Let $a$ be a world-altering action.
By $Res(a, M, \alpha)$ we denote the
Kripke structure $M'$ which is obtained from $M$ as
follows:
\begin{list}{$\bullet$}{\topsep=1pt \parsep=0pt \itemsep=1pt}
\item $M'[S] = \{r(a,u) \mid M[\pi](u) \models pre(a) \}$ where,
for each state $u \in M[S]$, $r(a,u)$ denotes a new and
distinguished state symbol;

\item $M'[\pi](r(a,u)) = res(a,u)$;

\item $(r(a,u), r(a,v)) \in M'[i]$ if $(u,v) \in M[i]$
and $r(a,u), r(a,v) \in M'[s]$ for $i \in \alpha$.
\end{list}
Let $(M,s)$ be a pointed Kripke structure and let $a$ be a world-altering
action.
The successor pointed Kripke structure of
$a$ in $(M,s)$ is defined as follows.
\begin{enumerate}
\item If $(M,s) \not\models pre(a)$ then $succ(a,(M,s))= \bot$.

\item If $(M,s) \not\models pre(a)$ then $succ(a,(M,s))$ is
the pointed Kripke structure $(M',s')$:
\begin{list}{$\bullet$}{\topsep=1pt \parsep=0pt \itemsep=1pt}
\item $M'[S] = Q[S]  \cup M[S]$;

\item $M'[\pi](u) = M[\pi](u)$ for $u \in M[S]$ and
      $M'[\pi](u) = Q[\pi](u)$ for $u \in Q[S]$;

\item $M'[i] = M[i] \cup Q[i] \cup R(i)$ where $R(i) = \emptyset$
for $i \in \alpha$ and
$R(i) = \{(r(a,u), v) \mid r(a,u) \in Q[S], v \in M[S], (u,v) \in M[i]\}$;
and

\item $s' = r(a,s)$.
\end{list}
where $Q = Res(a, M, \alpha)$,
and  $a \causes \varphi \iif \psi \performedby \alpha$ is the axiom
of $a$ applicable in $s$.
\end{enumerate}
This can be mapped to the following Prolog encoding:

{\small\begin{verbatim}
succ(sit(kripke(N,E),S), altering(If,Then,Alpha), sit(kripke(N1,E1),St1)) :-
    replica(kripke(N,E), kripke(N2,E2), Map), non_agents(Alpha,[],Gamma),
    remove_agent_edges(kripke(N2,E2), Gamma, kripke(N3,E3)),
    update_interpretations(N3,kripke(N3,E3),If, Then, N4),
    knowledge_union(kripke(N,E), kripke(N4,E3), Alpha, Map, kripke(N1,E1)),
    S = node(NameS,_), member([S1, NameS],Map),
    member(node(S1,Int1),N1), St1 = node(S1,Int1).
update_interpretations([],_,_,_,[]).
update_interpretations([node(Name,Inter)|Rest], K, If, Then, 
                               [node(Name,NewInter)|NewRest]) :-
    (hold(If,sit(K,node(Name,Inter))) -> 
        change_interpretation(Then,Inter,NewInter); NewInter = Inter),
    update_interpretations(Rest,K,If,Then,NewRest).
change_interpretation(F,I1,I2):- literal(F),!, single_update(F,I1,I2).
change_interpretation(and(F1,F2), I1, I2) :-
    change_interpretation(F1,I1,I3), change_interpretation(F2,I3,I2).
single_update(L, I1, I2):-  get_literal_fluent(L,F), delete(I1, value(F,_), I3),
    (fluent(L) -> I2 = [value(F,true)|I3]; I2 = [value(F,false)|I3]).
\end{verbatim}}

%%%%%%%%%%%%%%%%%%%%%%%%%%%%%%%%%%%%%%%%%%%%%%%%%%%%%%%%

\subsection{Query Answering and Planning in Prolog}
The previous encoding can be used to 
support hypothetical reasoning and planning in multi-agent theories. 
To answer queries of the form (\ref{query}) we can use the standard rules 
for computing the successor states of a sequence of action and verify 
the desirable properties in the final successor state. This is encoded 
as 

{\small
\begin{verbatim}
hold(Query,Seq):- initial(sit(kripke(Nodes,Edges),Node)), 
   succ(sit(kripke(Nodes,Edges),Node), Seq, S), hold(Query, S).  
succ(sit(kripke(Nodes,Edges),Node), [], sit(kripke(Nodes,Edges),Node)). 
succ(sit(kripke(Nodes,Edges),Node), [A|Seq], succ(S, Seq)):- 
   succ(sit(kripke(Nodes,Edges),Node), A, S).    
\end{verbatim}}

\noindent where $init(S)$ denotes that $S$ is the initial pointed Kripke structure and 
$hold(F,S)$ is the predicate that determines whether $F$ holds
w.r.t.  $S$. 

For planning, we need to determine a sequence
of actions that will accomplish a certain state of the 
world. The planning perspective supported by the implementation
below is that of an external observer, who has complete knowledge
about the capabilities of the different agents. 
A standard depth-first planner can be expressed using the 
following definition of the predicate {\tt depthplan}$(S,Plan,N)$---where 
$S$ is the initial pointed Kripke structure, $N$ is a bound on the 
maximal length of the plan, and $Plan$ is the actual sequence of 
actions.

{\small\begin{verbatim}
depthplan(S, Plan,N):-  depthplan(S,[],Plan,N).
depthplan(S, PlanIn, PlanIn,_):-  terminated(S).
depthplan(S0, PlanIn, PlanOut,N):- length(PlanIn,M), M < N, choose_action(A,S0), 
    build_action(A,Fmt), succ(S0,Fmt,S1),  depthplan(S1,[A|PlanIn],PlanOut,N).
choose_action(A,S) :- action(A,_), valid(A,S). 
\end{verbatim}}

The predicate {\tt build\_action} perform a
simple syntactic rearrangement of the action
representation (not reported).
The predicate {\tt valid} is used to 
ensure that the action is executable in 
the given pointed Kripke structure:

{\small 
\begin{verbatim} 
valid(A,S):- action(A,_Type), executability(A,Condition), hold(Condition,S).
\end{verbatim}
The test for termination ensures that the
goal has been achieved:
\begin{verbatim}
terminated(S) :- goal(Phi), hold(Phi,S).
\end{verbatim}}

As another example, it is easy to rewrite the
planner to perform a breadth-first exploration of
the search space.

{\small \begin{verbatim}
breadthplan(S,Plan,N):- breadthplan(S,Plan,0,N).
breadthplan(S,Plan,M,N):- M<N, plan(S,Plan,S1), length(Plan,M), terminated(S1).
breadthplan(S,Plan, M, N):- M < N, M1 is M+1, breadthplan(S,Plan,M1,N).
plan(S1,[],S1).
plan(S0,[A|B],S):- plan(S0,B,S1), build_action(A,Format), 
    valid(A,S1), succ(S1,Format,S).
\end{verbatim}}

Other forms of reasoning, e.g., switching from
the perspective of an external observer to the
perspective of an individual agent, can be similarly
encoded.

\section{Conclusion and Discussion}
\label{sec6}

In this paper, we investigated an application of logic
programming technology to the problem of manipulating Kripke
structures representing models of theories from the logic of 
knowledge. We illustrated how these foundations can be used
to provide a computational background for a novel action
language, $\mA$, to encode multi-agent planning domains where
agents can perform both world-altering actions as well as
actions affecting agents' knowledge. 

To the best of our knowledge, the 
encoding presented in this paper is the first implementation using Prolog of
an action language with such a set of features. The use of logic 
programming allows a very natural encoding of the semantics of 
$\mA$, and it facilitates the development of meta-interpreters 
implementing different forms of reasoning (e.g., observer-based
planning). Let us underline that  generic frameworks for reasoning with
modal logics have been proposed (e.g.,~\cite{Horrocks,Farinas}), and these
could be used as external solvers to address the specific tasks of recomputing
models---we will explore this option in our future work. Nevertheless, we
believe the updates of Kripke structures directly performed in LP and the
ability of embedding such process in a LP language, which offers other features
(e.g., constraint solving) makes LP a more suitable avenue for implementing 
languages like $\mA$. Although the prototype has not been subjected to formal testing
(we are working on identifying interesting domains from
the literature), simple planning tasks (e.g., develop a plan
that enables agents $a,c$ to learn about the status of the coin
while $b$ maintains its ignorance) can be solved within a few
seconds of computation. A formal experimental evaluation will be
part of the future work.

\bibliographystyle{acmtrans}
\end{document}